\documentclass[letterpaper]{article} 
\usepackage{aaai25}  
\usepackage{times}  
\usepackage{helvet}  
\usepackage{courier}  
\usepackage[hyphens]{url}  
\usepackage{graphicx} 
\usepackage{natbib}  
\usepackage{caption} 
\usepackage{algorithm}
\usepackage{algorithmic}

\usepackage{amsthm,amsmath,amssymb}
\usepackage{mathrsfs}
\usepackage{multirow}
\usepackage{booktabs}
\usepackage{arydshln}
\usepackage{pdfpages}
\usepackage{xcolor}
\definecolor{mygray}{gray}{0.6}

\usepackage{newfloat}
\usepackage{listings}
\DeclareCaptionStyle{ruled}{labelfont=normalfont,labelsep=colon,strut=off} 
\floatstyle{ruled}
\newfloat{listing}{tb}{lst}{}
\floatname{listing}{Listing}
\title{3D$^2$-Actor: Learning Pose-Conditioned 3D-Aware Denoiser for Realistic Gaussian Avatar Modeling}
\author{
    Zichen Tang\textsuperscript{\rm 1},
    Hongyu Yang\textsuperscript{\rm 1,2}\thanks{Corresponding author.},
    Hanchen Zhang\textsuperscript{\rm 3},
    Jiaxin Chen\textsuperscript{\rm 3},
    Di Huang\textsuperscript{\rm 3}
}
\affiliations{
    \textsuperscript{\rm 1}School of Artificial Intelligence, Beihang University, Beijing, China\\
    \textsuperscript{\rm 2}Shanghai Artificial Intelligence Laboratory, Shanghai, China\\
    \textsuperscript{\rm 3}School of Computer Science and Engineering, Beihang University, Beijing, China\\
    \{zctang, hongyuyang, zhanghanchen, jiaxinchen, dhuang\}@buaa.edu.cn
}

\begin{document}

\maketitle

\begin{abstract}
Advancements in neural implicit representations and differentiable rendering have markedly improved the ability to learn animatable 3D avatars from sparse multi-view RGB videos. However, current methods that map observation space to canonical space often face challenges in capturing pose-dependent details and generalizing to novel poses. While diffusion models have demonstrated remarkable zero-shot capabilities in 2D image generation, their potential for creating animatable 3D avatars from 2D inputs remains underexplored. In this work, we introduce 3D$^2$-Actor, a novel approach featuring a pose-conditioned 3D-aware human modeling pipeline that integrates iterative 2D denoising and 3D rectifying steps. The 2D denoiser, guided by pose cues, generates detailed multi-view images that provide the rich feature set necessary for high-fidelity 3D reconstruction and pose rendering. Complementing this, our Gaussian-based 3D rectifier renders images with enhanced 3D consistency through a two-stage projection strategy and a novel local coordinate representation. Additionally, we propose an innovative sampling strategy to ensure smooth temporal continuity across frames in video synthesis. Our method effectively addresses the limitations of traditional numerical solutions in handling ill-posed mappings, producing realistic and animatable 3D human avatars. Experimental results demonstrate that 3D$^2$-Actor excels in high-fidelity avatar modeling and robustly generalizes to novel poses. Code is available at: https://github.com/silence-tang/GaussianActor.
\end{abstract}

%

\section{Introduction}

Reconstructing animatable 3D human avatars is essential for applications in VR/AR, the Metaverse, and gaming. However, the task is challenging due to factors like non-rigid complex motions and the stochastic nature of subtle clothing wrinkles, which complicate realistic human actor modeling.

Traditional methods \cite{collet2015high, dou2016fusion4d, bogo2015detailed, shapiro2014rapid} are often hindered by labor-intensive manual design and the difficulty of acquiring high-quality data, limiting their applicability in real-world scenarios. Recent advances in neural implicit representations and differentiable neural rendering \cite{sitzmann2020implicit, wang2021neus, lombardi2019neural, gao2021dynamic, park2021hypernerf, pumarola2021d} have opened new avenues for character reconstruction and animation from sparse multi-view RGB videos. While techniques like Neural Radiance Field (NeRF) \cite{mildenhall2021nerf} excel in synthesizing static scenes, achieving high-fidelity results for dynamic human avatars remains a significant challenge.

One prominent approach involves using a deformation field to map the observation space to a canonical space, as demonstrated in methods like \cite{su2021nerf, peng2021animatable}. Although learning this backward mapping is relatively straightforward, its generalization to novel poses is often limited due to its reliance on the observation state. Alternatives that employ forward mapping \cite{wang2022arah, li2022tava}, utilizing techniques like differentiable root-finding, have been proposed to address these generalization challenges. Additionally, Neural Body \cite{peng2021neural} introduces a conditional NeRF approach that anchors local features on SMPL \cite{loper2015smpl} vertices, which serve as a scaffold for the model.

Despite these advancements, current methods face limitations in handling the complex dynamics of human bodies. For instance, the high stochasticity of clothing, characterized by delicate wrinkles that appear and disappear, poses a significant challenge. Approaches such as \cite{peng2021neural, peng2021animatable, wang2022arah} optimize per-frame latent codes to capture this variability but struggle to adapt to novel poses due to the limited expressivity of these latent codes.
More recent works, such as \cite{hu2023gauhuman, qian2023gaussianavatars, li2024animatable}, have integrated 3D Gaussian Splatting (3DGS) \cite{kerbl20233d} into their pipelines, delivering significantly improved results in terms of both rendering efficiency and fidelity. However, these approaches do not fully consider finer visual details during novel pose synthesis.

In contrast to these 3D-based approaches, recent 2D generative diffusion models \cite{ho2020denoising, rombach2022high} have demonstrated significant advantages in terms of visual quality. However, the absence of a 3D representation presents a challenge when extending 2D diffusion models to maintain spatial and temporal consistency, particularly in human-centric scenarios. Several works have attempted to achieve 3D-consistent generation by incorporating additional control inputs \cite{liu2023zero} or integrating a 3D representation into the workflow \cite{liu2024syncdreamer, anciukevivcius2023renderdiffusion, karnewar2023holodiffusion}. These methods mainly focus on single-scene generation for general objects, often neglecting temporal consistency, which makes them inadequate for dynamic human modeling.

In this work, we tackle the challenge of reconstructing and animating high-fidelity 3D human avatars with controllable poses by introducing \textbf{3D$^2$-Actor}, a novel approach featuring a 3D-aware denoiser composed of interleaved pose-conditioned 2D denoising and 3D rectifying steps. Our method uniquely combines the strengths of 3DGS and 2D diffusion models to achieve superior performance in human-centric tasks. Specifically, the 2D denoiser is conditioned on pose clues to generate detailed multi-view images, which are essential for providing rich features supporting the following high-fidelity 3D reconstruction and rendering process. Additionally, the 2D denoiser enhances intricate details from preceding 2D or 3D steps, thereby improving the overall fidelity of the avatar. Complementing the 2D denoiser, our 3D rectifier employs a novel two-stage projection strategy combined with a mesh-based local coordinate representation. The rectifier queries positional offsets and other 3D Gaussian attributes from input images to produce structurally refined multi-view renderings. The integration of 3D Gaussian Splatting ensures high morphological integrity and consistent 3D modeling across various views. To address the temporal incontinuity in animated avatar videos, we propose a Gaussian consistency sampling strategy. This technique utilizes Gaussian local coordinates from previous frames to determine current positions, enabling smooth inter-frame transitions without the need for additional temporal smoothing modules. Our key contributions include:
\begin{itemize}
\item Novel 3D-Aware Denoiser: We propose a 3D-aware denoiser tailored for reconstructing animatable human avatars from multi-view RGB videos. This method integrates the generative capabilities of 2D diffusion models with the efficient rendering of 3D Gaussian Splatting.
\item Advanced 3D Rectifier: Our 3D rectifier incorporates a two-stage projection module and a novel local coordinate representation to render structurally refined frames with high multi-view consistency. 
\item Gaussian Consistency Sampling Strategy: We propose a simple yet effective sampling strategy that ensures inter-frame continuity in generated avatar videos. This approach preserves temporal consistency and enhances the overall quality of animated sequences.
\end{itemize}

\section{Related Work}

\subsection{Animatable 3D Human Avatars}

In recent years, significant advancements in neural scene representations and differentiable neural rendering techniques have demonstrated high effectiveness in synthesizing novel views for both static \cite{mildenhall2021nerf, sitzmann2020implicit} and dynamic scenes \cite{gao2021dynamic, park2021hypernerf, pumarola2021d}. Building upon these studies, various methods attempt to realize 3D human \textbf{reconstruction} from sparse-view RGB videos. 

Among these approaches, a common line of works involve learning a backward mapping to project points from the observation space to the canonical space. A-Nerf \cite{su2021nerf} constructs a deterministic backward mapping using bone-relative embeddings. Animatable NeRF \cite{peng2021animatable} trains a backward LBS network, yet it encounters challenges in generalizing to poses beyond the distribution. ARAH \cite{wang2022arah} and TAVA \cite{li2022tava}, in contrast, utilize a forward mapping to transfer features from the canonical space to the observation space. While the generalizability to novel poses has been improved by these methods, the computational cost of their differentiable root-finding algorithm is quite high.

\begin{figure*}[t]
  \centering
  \includegraphics[width=1.0\textwidth]{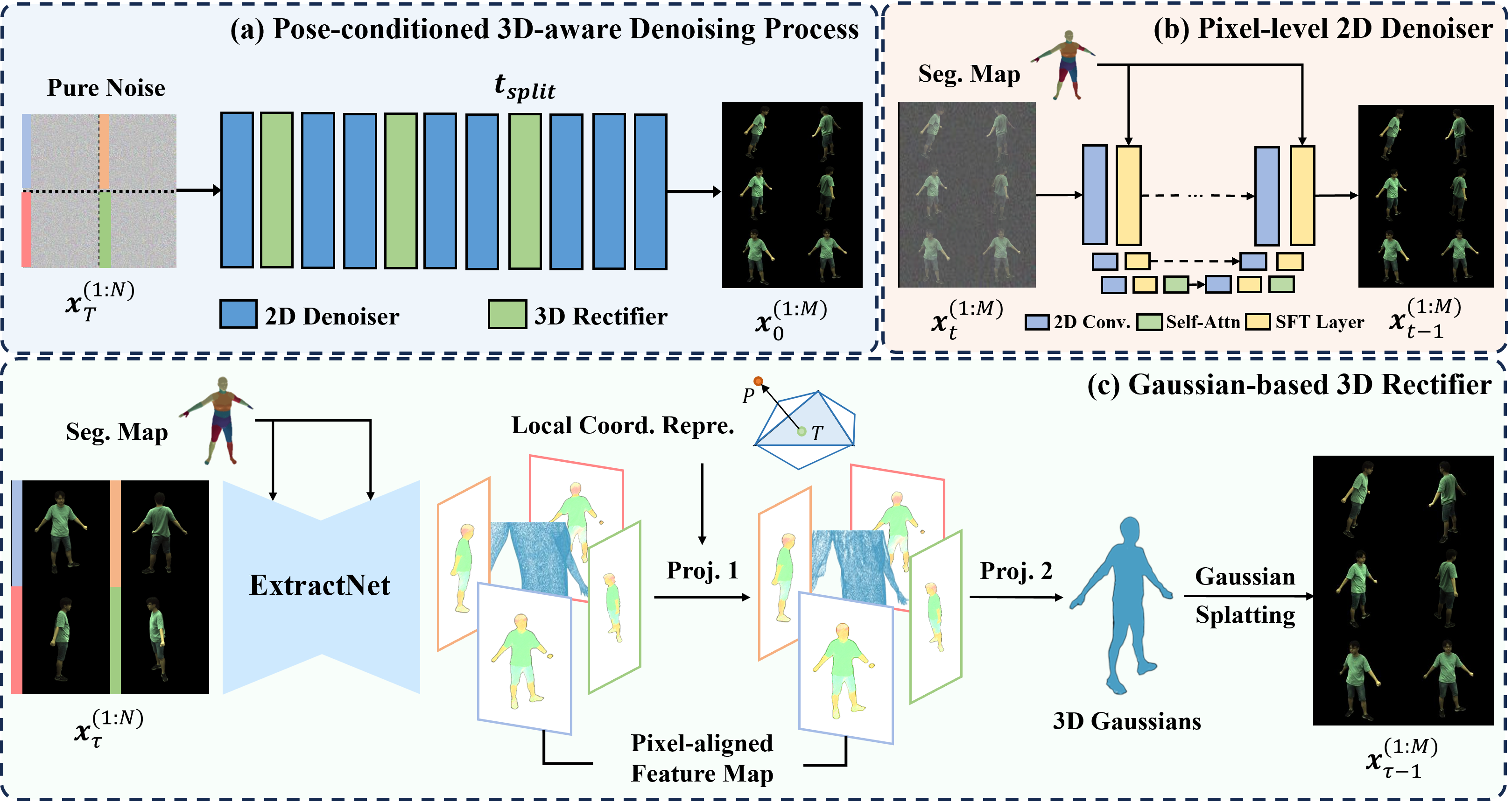}
  \caption{Illustration of the 3D-aware denoising process. (a) The 3D-aware denoising pipeline consists of interlaced 2D and 3D steps. It begins with pure noise input, progressively generating realistic multi-view images of the human avatar with the assistance of pose information. (b) Guided by body segmentation maps as pose cues, the 2D denoiser (blue box in (a)) transforms noised images from the previous 2D or 3D steps into clean ones with enhanced intricate details. It also provides clean images for the subsequent 3D rectifier to achieve accurate 3D human avatar modeling. (c) Given clean images from \(N\) anchor views, the 3D rectifier (green box in (a)) performs a two-stage projection leveraging a mesh-based Gaussian local coordinate representation to reconstruct 3D Gaussians, enabling the rendering of multi-view human images with high 3D consistency.} 
  \label{fig:pipeline}
\end{figure*}

Another line of works focus on creating a conditional NeRF for modeling dynamic human bodies. Neural Body \cite{peng2021neural} attaches structured latent codes to posed SMPL vertices and diffuses them into the adjacent 3D space. Despite its capability for high-quality view synthesis, this method performs suboptimally with novel poses. NPC \cite{su2023iccv} employs points to store high-frequency details and utilizes a graph neural network to model pose-dependent deformation based on skeleton poses.

A key focus of human avatar \textbf{animation} lies in how to transform input poses into changes in appearance. PoseVocab \cite{li2023posevocab} proposes joint-structured pose embeddings to encode dynamic human appearance, successfully mapping low-frequency SMPL-derived attributes to high-frequency dynamic human appearances. However, it neglects the fact that identical poses in different motions can result in varying appearances. Some methods \cite{peng2021neural, peng2021animatable, wang2022arah} employ a per-frame global latent vector to encode stochastic information but this representation cannot generalize well to novel poses. In contrast, our method directly models the distribution of appearances under various poses in image space, enabling a more effective capture of high-frequency visual details.

The advent of 3D Gaussian splatting \cite{kerbl20233d} has unlocked new possibilities for high-fidelity avatar reconstruction with real-time rendering. A number of concurrent methods \cite{hu2023gauhuman, jung2023deformable, li2023human101, qian2023gaussianavatars, qian20233dgs, liu2024gea} have investigated the integration of 3D Gaussian with SMPL models for constructing a 3D Gaussian avatar. While the majority of them try to improve rendering efficiency by substituting the neural implicit radiance field with 3D Gaussian representation, our work focuses more on improving the modeling of detailed appearances to enhance image quality.

\subsection{3D Diffusion Models}

Diffusion models \cite{ho2020denoising} have demonstrated superior performance in 2D image generation. Due to the absence of a standardized 3D data representation, the expansion of 2D diffusion models into the 3D domain remains an unresolved issue. Some studies \cite{nichol2022point, gupta20233dgen, muller2023diffrf} employ 3D supervision to achieve direct generation of 3D content. However, their practical effectiveness is constrained by the limited size and diversity of the available training data \cite{po2023state}.

Inspired by 3D GANs, various approaches \cite{anciukevivcius2023renderdiffusion,  karnewar2023holodiffusion, ssdnerf, Szymanowicz_2023_ICCV} have been proposed to directly train a diffusion model using 2D image datasets. RenderDiffusion \cite{anciukevivcius2023renderdiffusion} builds a 3D-aware denoiser by incorporating tri-plane representation and neural rendering, predicting a clean image from the noised 2D image. Building upon this research, Viewset Diffusion \cite{Szymanowicz_2023_ICCV} extends it to multi-view settings. In contrast to these approaches, we take a step further to achieve pose-conditioned human-centric generation. We also present a meticulously designed sampling strategy, enabling the smooth generation of dynamic human videos without introducing extra temporal modules, which is not achieved by current works.


\section{Method}

\textbf{Problem statement.} Given multi-view RGB videos of a single human actor as training data, a model should be trained to reconstruct a realistic 3D avatar of the actor and generate high-fidelity and temporal-smoothing videos when performing avatar animation. In the following sections, preliminary will be present first. Next, we will introduce our 3D-aware denoising process combined with 2D denoiser and 3D rectifier. Following that, a simple yet effective inter-frame sampling strategy will be detailed. Finally, we will elucidate the training objectives of our proposed 3D and 2D modules.

\subsection{Preliminary}

\textbf{Diffusion and denoising process} is the core of diffusion models. In this work, we extend this process to our multi-view setting, data $\boldsymbol{x}:=\boldsymbol{x}^{(1:N)}$ represents a set of $N$ images that consistently depict a 3D human avatar. To establish the correlation between the noise distribution and the data distribution, a hierarchy of variables is defined as $\boldsymbol{x}_t^{(1:N)}$, $t=0,\dots,T$, where $\boldsymbol{x}_T^{(1:N)}\sim\mathcal{N}(0, \mathbf{I})$ and $\boldsymbol{x}_0^{(1:N)}$ is the set of generated multi-view images.
Leveraging the properties of the Gaussian distribution, the forward diffusion process that gradually introduces Gaussian noises to clean data $\boldsymbol{x}_0^{(1:N)}$ can be rewritten as:
\begin{equation}\label{eq:0}
    q(\boldsymbol{x}_t^{(1:N)} | \boldsymbol{x}_0^{(1:N)}) =
    \mathcal{N}(\boldsymbol{x}_t^{(1:N)}; \sqrt{\bar\alpha_t} \boldsymbol{x}_0^{(1:N)}, (1-\bar\alpha_t)\mathbf{I}),
\end{equation}
where $\bar\alpha_t = \prod_{i=1}^{t} \alpha_i$ and $\alpha_i$ denotes the predefined schedule constant. Correspondingly, the inverse process can be formulated as:
\begin{equation}\label{eq:1}
    p(\boldsymbol{x}_{t-1}^{(1:N)} | \boldsymbol{x}_{t}^{(1:N)}) = 
    \mathcal{N}(\boldsymbol{x}_{t-1}^{(1:N)}; \mu_\theta(\boldsymbol{x}_t^{(1:N)}, t), \Sigma_\theta(\boldsymbol{x}_t^{(1:N)}, t)),
\end{equation}
where the mean and variance can be estimated through a U-Net $D_\theta$ trained with loss $L$ to reconstruct the clean data $x_0^{(1:N)}$ from the noised counterpart $x_t^{(1:N)}$:
\begin{equation}\label{eq:2}
    L=\|D_\theta(\boldsymbol{x}_t^{(1:N)}, t) - \boldsymbol{x}_0^{(1:N)}\|^2.
\end{equation}

\noindent \textbf{3D Gaussian Splatting} \cite{kerbl20233d} is an effective point-based representation consisting of a set of anisotropic Gaussians. Each 3D Gaussian is parameterized by its center position $\boldsymbol{\mu} \in \mathbb{R}^3$, covariance matrix $\boldsymbol{\Sigma} \in \mathbb{R}^7$, opacity $\alpha \in \mathbb{R}$ and color $\boldsymbol{c} \in \mathbb{R}^3$. By splatting 3D Gaussians onto 2D image planes, we can perform point-based rendering:

\begin{equation}
\begin{gathered}
G(\boldsymbol{p}, \boldsymbol{\mu}_{i}, \boldsymbol{\Sigma}_{i}) = \exp(- \frac{1}{2}(\boldsymbol{p} - \boldsymbol{\mu}_{i})^\intercal \boldsymbol{\Sigma}_{i}^{-1}(\boldsymbol{p} - \boldsymbol{\mu}_{i})), \\
\boldsymbol{c}(\boldsymbol{p}) = \sum\limits_{i\in \mathcal{K}} \boldsymbol{c}_{i}\alpha_{i}' \prod\limits_{j=1}^{i-1} (1 - \alpha_{j}'), \alpha_{i}' = \alpha_{i} G(\boldsymbol{p}, \boldsymbol{\mu}_{i}, \boldsymbol{\Sigma}_{i}).
\end{gathered}
\end{equation}

\noindent Here, $\boldsymbol{p}$ is the coordinate of the queried point. $\boldsymbol{\mu}_i$, $\boldsymbol{\Sigma}_i$, $\boldsymbol{c}_i$, $\alpha_i$, and $\alpha_i'$ denote the center, covariance, color, opacity, and density of the $i$-th Gaussian, respectively. $G(\boldsymbol{p}, \boldsymbol{\mu}_i, \boldsymbol{\Sigma}_i)$ represents the value of the $i$-th Gaussian at position $\boldsymbol{p}$. $\mathcal{K}$ is a sorted list of Gaussians in this tile.

\subsection{3D-aware Denoising Process}

To facilitate human avatar reconstruction (given seen poses) and animation (given novel poses), we innovatively propose a generative 3D-aware denoising process which takes as input pure noise from $N$ anchor views and SMPL \cite{loper2015smpl} pose information and outputs high-quality clean images of the clothed human body. Due to the fact that the 2D denoiser denoises images from different views independently at each step, it fails to ensure consistency in human geometry and texture across views. To address this issue, $k$ 3D rectifying steps are inserted between the 2D denoising steps to maintain the 3D consistency of generated images. Considering that the overall denoising process generates large-scale global structure at early stages and finer details at later stages \cite{Huang2023_dreamtime}, and that 3D consistency among multi-view images is mainly reflected in large-scale features, we merely insert 3D steps in the early stages. This approach aims to improve 3D consistency without jeopardizing the quality of fine texture generation. As illustrated in Fig. \ref{fig:pipeline}, we first apply the initial 3D rectifying step after the first 2D denoising step. Then, we select a timestep $t_{split}$ as the split point between the early and later stages of denoising and insert the final 3D rectifying step at this point. Subsequently, $k-2$ 3D rectifying steps are evenly inserted between these 2D steps. Finally, a few 2D steps are appended to the overall denoising process, further optimizing the local delicate textures. The details of the the 2D denoiser and the 3D rectifier will be introduced below.

\subsection{Pixel-level 2D Denoiser}

The 2D denoiser is a fundamental component of our 3D-aware denoising process. Basically, it functions as a refiner that enhances local details in the output images from prior 2D or 3D steps. It can also provide clean images for the subsequent 3D rectifying step. Our 2D denoiser acts like a U-Net \cite{ronneberger2015u}, taking noisy images, human body segmentation maps and the denoising timestep $t$ as inputs to predict the denoised clean image at each step. To effectively incorporate pose cues, we draw inspiration from SFTGAN \cite{wang2018sftgan} and introduce an SFT layer into each U-Net block to modulate the output of the 2D convolution layer. Given that the 3D rectifier can render images from any camera view, our 2D denoiser is trained on frames with varying views to ensure robustness.

\subsection{Gaussian-based 3D Rectifier}

The 3D rectifier plays an essential role in our 3D-aware denoising process. It takes in clean images $\boldsymbol{I}^{(1:N)}$ from $N$ anchor views produced by the previous 2D denoiser and reconstructs the current 3D Gaussians of the avatar. Then, real-time rendering of structure-aligned multi-view images with higher 3D consistency (than the previous 2D step) can be achieved. Note that the 3D rectifier outputs clean images, which can be regarded as ``$\boldsymbol{x}_0$''. Therefore, we can naturally integrate it with the next denoising step leveraging the DDIM \cite{song2020denoising} sampling trick:

\begin{equation}
    \boldsymbol{x}_{t-1}^{(1:N)} = \sqrt{\alpha_{t-1}} \hat{\boldsymbol{x}}_0^{(1:N)} + c_t \hat{\epsilon}_t^{(1:N)} + \sigma_t \epsilon_t^{(1:N)},
\end{equation}
where $c_t=\sqrt{1-\alpha_{t-1}-\sigma_t^2}$ and $\sigma_t$ are necessary coefficients, $\hat{\boldsymbol{x}}_0^{(1:N)}$ is the output of the 3D rectifier and $\epsilon_t^{(1:N)}$ is sampled random noise.

Specifically, we start by rendering body segmentation maps $\boldsymbol{S}^{(1:N)}=\mathcal{R}(\mathcal{M}, \boldsymbol{c}^{(1:N)})$, where $\mathcal{M}$ and $\boldsymbol{c}^{(1:N)}$ denote the current posed SMPL model and camera poses, respectively, and $\mathcal{R}$ is the mesh rasterizer. They serve as pose conditions to aid the neural network $f_{ext}$ in feature extraction for perceiving the 3D actor. Similar to the 2D denoiser, we also insert SFT layers into each U-Net block, effectively leveraging the pose guidance. The entire process of extracting pixel-aligned features can be formulated as:
\begin{equation}\label{eq:encoder}
    \boldsymbol{F}_{pix}^{(1:N)} = f_{ext}(\boldsymbol{I}^{(1:N)}, \boldsymbol{S}^{(1:N)}, t).
\end{equation}

\begin{figure*}[t]
  \centering
  \includegraphics[width=1.0\textwidth]{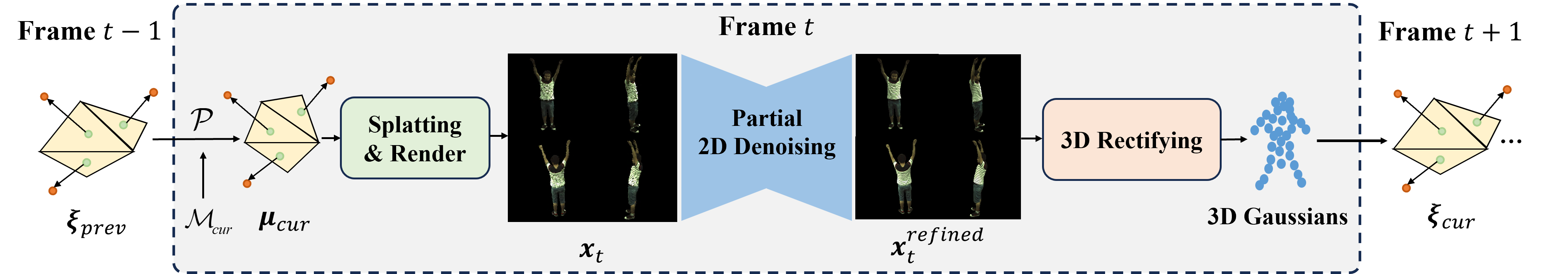}
  \caption{An illustration of the inter-frame Gaussian consistency sampling strategy for improving temporal continuity.}
  \label{fig:sampling}
\end{figure*}

After pixel-aligned features are fetched, a key question is how to build the 3D representation of the avatar. Considering the flexibility and efficiency of 3D Gaussian Splatting, we choose it as our 3D representation. Different from current works \cite{li2024animatable, jiang2024uv} which use regressed 2D maps to store Gaussian attributes, we seek to predict these attributes with a two-stage projection strategy which fully exploits the 3D spatial information. 

\textbf{Stage 1: query Gaussian local coordinates.} For any Gaussian $P$ in the 3D space, its projection $T$ on the nearest triangle mesh can be represented by barycentric coordinates $(\lambda_1, \lambda_2, \lambda_3)$, where $\lambda_1 + \lambda_2 + \lambda_3 = 1$. Therefore, $P$ can be easily described by a local coordinate quaternion $\boldsymbol{\xi} = (\lambda_1, \lambda_2, \lambda_3, m)$, where $m=|\boldsymbol{TP}|$ and we can derive the actual position of $P$ by $P = \lambda_1A + \lambda_2B + \lambda_3C + m\boldsymbol{n}$, where $A, B, C$ and $\boldsymbol{n}$ are the vertex positions and the normal vector of the triangle mesh, respectively. After Gaussian positions are initialized by sampling uniformly on the SMPL mesh, we project each Gaussian onto $\boldsymbol{F}_{pix}^{(1:N)}$ to query their position displacements. Rather than directly querying their displacements in observation space, we choose to query their local coordinates instead. This trick constrains the Gaussian movement within a reasonable range, helps model subtle clothes wrinkles and facilitates our inter-frame sampling strategy. Specifically, We project each Gaussian onto $\boldsymbol{F}_{pix}^{(1:N)}$, apply bilinear interpolation to obtain feature vectors for each view, and subsequently concatenate them along the feature dimension. If a Gaussian is not visible under a certain view, the corresponding vector is set to zero. Afterwards, a light-weight MLP takes Gaussian positions in the canonical pose space and the previously obtained projected features as input to predict Gaussian local coordinates. \textbf{Stage 2: query other Gaussian attributes.} After deriving actual Gaussian positions with local coordinates, we apply another projection to fetch the offsets of the remaining Gaussian attributes. Finally, clean multi-view images with higher 3D consistency can be rendered rapidly.

\subsection{Inter-frame Gaussian Consistency Sampling}

\label{sec:IGC sampling}

From the previous discussion, we know that the 3D-aware denoising process can generate highly realistic single-frame renderings from pure noise. However, applying this independently to each frame in video generation can cause noticeable inconsistencies, severely affecting the visual quality. Adding a temporal module is a possible way to address this issue, but this comes with an increased computational cost for training. In contrast, we design a novel inter-frame Gaussian consistency sampling strategy during inference to ensure seamless inter-frame transitions when synthesizing videos. The core idea is to use information from the previous frame to generate a rough image of the current frame, then perform several late-stage 2D denoising steps to correct visual artifacts. To obtain the current frame's Gaussians, we should propagate the SMPL pose change to the change of Gaussian positions. Fortunately, this can be achieved easily using our mesh-based local coordinates. As depicted in Fig. \ref{fig:sampling}, given Gaussian local coordinates $\boldsymbol{\xi}_{prev}^{(1:n)}$ of the last frame and the current frame's SMPL mesh $\mathcal{M}_{cur}$, the current Gaussian positions $\boldsymbol{\mu}_{cur}^{(1:n)}$ can be derived by:

\begin{equation}\label{eq:gs_trans}
    \boldsymbol{\mu}_{cur}^{(1:n)} = \mathcal{P}(\boldsymbol{\xi}_{prev}^{(1:n)}, \mathcal{M}_{cur}),
\end{equation}
where $n$ is the number of Gaussians and $\mathcal{P}$ is an operation that transforms Gaussian local coordinates to their actual positions in the observation space. However, rendering images directly using these Gaussians may lead to noticeable artifacts. To mitigate this, we add slight noise to the rendered images and perform several 2D denoising steps from a smaller timestep, yielding more plausible results. Finally, we apply an additional 3D step to obtain the 3D Gaussians of the current frame. Adopting this sampling strategy for video generation offers distinct advantages in terms of inter-frame continuity compared to generating each frame separately. It also provides computational efficiency, as denoising is only performed partially from a relatively small timestep.

\subsection{Training objective}

The complete training process includes two separate training workflows for the 3D rectifier $G_{3D}$ and the 2D denoiser $D_{2D}$. $G_{3D}$ is trained with a loss function that includes both photometric loss and mask loss. Given clean video frames $\boldsymbol{I}_f^{(1:N)}$ from $N$ anchor views and conditional SMPL segmentation maps $\boldsymbol{S}_f^{(1:N)}$ of frame $f$, the training objective is to reconstruct accurate 3D Gaussians from the given input to achieve consistent 3D rendering from $M$ specified views $\boldsymbol{c}^{(1:M)}$. The loss function measures the similarity between the rendered multi-view images and the ground-truth images, including the $L_2$ loss for the RGB images:

\begin{equation}\label{eq:loss_3d_rgb}
     L_{rgb} = \|G_{3D}(\boldsymbol{I}_{f}^{(1:N)}, \boldsymbol{S}_{f}^{(1:N)}, \boldsymbol{c}^{(1:M)}) - \boldsymbol{I}_{f}^{(1:M)} \|^2,
\end{equation}
and the $L_2$ loss $L_{mask}$ for the masks, which is omitted for brevity. The overall loss of $G_{3D}$ can be represented as:

\begin{equation}\label{eq:loss_3d_mask}
     L_{3D} = \lambda_{rgb} L_{rgb} + \lambda_{mask} L_{mask}.
\end{equation}

In terms of the 2D denoiser $D_{2D}$, given a clean video frame $\boldsymbol{I}_f$, the corresponding SMPL segmentation map $\boldsymbol{S}_f$ and timestep $t$, we only apply RGB loss to train the model:

\begin{equation}\label{eq:loss_2d_rgb}
     L_{2D} = \|D_{2D}(\boldsymbol{I}_f, \boldsymbol{S}_f, t) - \boldsymbol{I}_f \|^2.
\end{equation}


\begin{figure}[t]
  \centering
  \includegraphics[width=1.0\columnwidth]{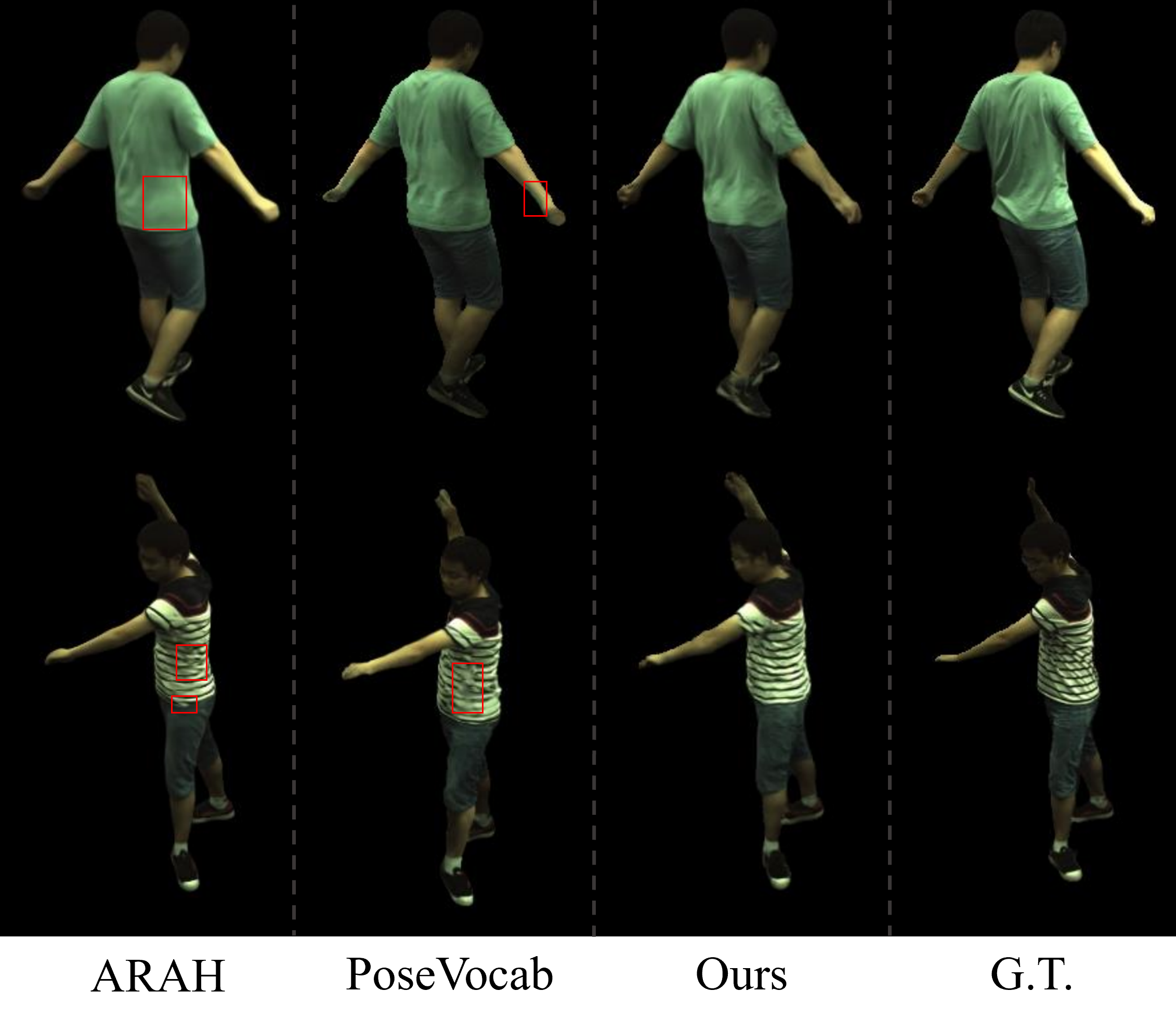}
  \caption{Qualitative comparison of single-frame novel pose synthesis results against ARAH \cite{wang2022arah} and PoseVocab \cite{li2023posevocab} on sequences 313 and 315 of the ZJU-MoCap dataset. Please zoom in for better observation.}
  \label{fig:single_frame}
\end{figure}

\begin{table*}[tb]

    \centering
    
    \begin{tabular}{@{}cccc@{\hspace{2pt}}ccc@{\hspace{2pt}}ccc@{\hspace{2pt}}ccc@{}}
    
    \toprule
  
    \multirow{2}{*}{Method} & \multicolumn{3}{c}{313} & \multicolumn{3}{c}{315} & \multicolumn{3}{c}{377} & \multicolumn{3}{c}{386}
  
    \\ \cmidrule(l){2-13} & {\textcolor{mygray}{PSNR$\uparrow$}} & {\hspace{-10pt} LPIPS$\downarrow$} & {\hspace{-7pt}FID$\downarrow$} & {\textcolor{mygray}{PSNR$\uparrow$}} & {\hspace{-7pt}LPIPS$\downarrow$} & {\hspace{-7pt}FID$\downarrow$} & {\textcolor{mygray}{PSNR$\uparrow$}} & {\hspace{-7pt}LPIPS$\downarrow$} & {\hspace{-7pt}FID$\downarrow$} & {\textcolor{mygray}{PSNR$\uparrow$}} & {\hspace{-7pt}LPIPS$\downarrow$} & {\hspace{-7pt}FID$\downarrow$} 
  
    \\ \midrule
    
    ARAH & \textbf{\textcolor{mygray}{24.3}} & \hspace{-12pt} 0.097 & \hspace{-10pt} 32.6 & \textbf{\textcolor{mygray}{20.9}} & \hspace{-12pt} 0.102 & \hspace{-10pt} 31.4 & \textbf{\textcolor{mygray}{24.8}} & \hspace{-12pt} 0.109 & \hspace{-10pt} 32.6 & \textbf{\textcolor{mygray}{27.9}} & \hspace{-12pt} 0.152 & \hspace{-10pt} 54.6  \\
  
    PoseVocab & \textcolor{mygray}{23.3} & \hspace{-12pt} 0.101 & \hspace{-10pt} 27.5 & \textcolor{mygray}{20.6} & \hspace{-12pt} 0.100 & \hspace{-10pt} 27.2 & \textcolor{mygray}{24.1} & \hspace{-12pt} 0.091 & \hspace{-10pt} \textbf{25.8} & \underline{\textcolor{mygray}{26.8}} & \hspace{-12pt} 0.134 & \hspace{-10pt} \underline{31.9} \\
  
    Ours & \underline{\textcolor{mygray}{23.5}} & \hspace{-12pt} \textbf{0.080} & \hspace{-10pt} \textbf{19.5} & \underline{\textcolor{mygray}{20.7}} & \hspace{-12pt} \textbf{0.090} & \hspace{-10pt} \textbf{20.2} & \underline{\textcolor{mygray}{24.4}} & \hspace{-12pt} \textbf{0.090} & \hspace{-10pt} \underline{26.4} & \underline{\textcolor{mygray}{26.8}} & \hspace{-12pt} \textbf{0.123} & \hspace{-10pt} \textbf{28.6}
    
    \\ \bottomrule
  
    \end{tabular}

    \caption{Quantitative comparison of single-frame novel pose synthesis against ARAH \cite{wang2022arah} and PoseVocab \cite{li2023posevocab} on 4 sequences of the ZJU-MoCap dataset. Bold indicates the best, while underline denotes the second-best.}
    \label{tab:single_frame}
    
\end{table*}

\section{Experiments}

\subsection{Implementation Details}
The 3D rectifier takes clean images at a resolution of 512$\times$512 from $N=4$ anchor views as input, reconstructs 3D avatar Gaussians, and renders $M=8$ multi-view images at the same resolution. The number of Gaussians sampled on the SMPL \cite{loper2015smpl} mesh is $n=373056$. We train this model with a learning rate of $5\times10^{-5}$. The 2D denoiser functions at a resolution of 512$\times$512, consistent with the resolution of the ground-truth images. This model is trained with a learning rate of $4\times10^{-4}$. When conducting single frame novel pose synthesis, our 3D-aware denoising process has 20 denoising steps in total. All experiments are conducted on NVIDIA RTX 3080 Ti GPUs.

\textbf{Dataset.} Our experiments are conducted on the ZJU-MoCap \cite{peng2021neural} dataset, which includes 9 sequences captured with 23 calibrated cameras. Each sequence features a video of an individual performing a specific action. We utilize 80\% of frames from each sequence for training and left the remaining frames for testing.

\textbf{Metrics.} We adopt Peak Signal-to-Noise Ratio (PSNR), Learned Perceptual Image Patch Similarity (LPIPS) \cite{Zhang_2018_CVPR}, and Frechet Inception Distance (FID) \cite{heusel2017gans} for quantitative evaluation.

\begin{figure}[t]
  \centering
  \includegraphics[width=1.0\columnwidth]{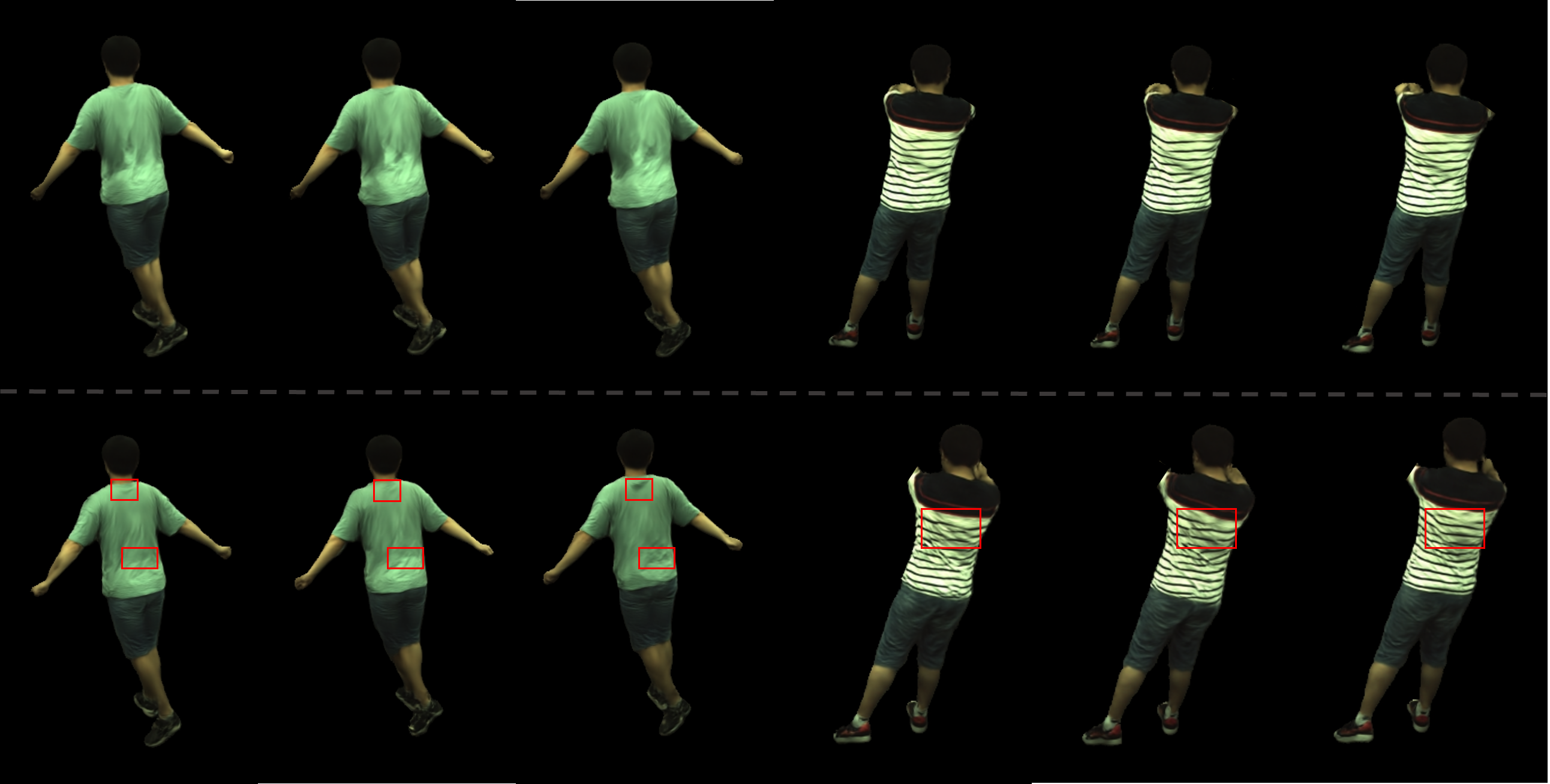}
  \caption{Consecutive frame generation results. Top row shows results using the proposed sampling strategy; bottom row displays results from independent sampling.}
  \label{fig:abl_sampling_strategy}
\end{figure}

\textbf{Baselines.} We compare our method with two state-of-the-art counterparts suitable for ZJU-MoCap dataset: ARAH \cite{wang2022arah} and PoseVocab \cite{li2023posevocab}. We retrained the baseline methods using their officially released code to align the training/test set split for all the methods. 

\subsection{Single-Frame Novel Pose Synthesis}

To evaluate the visual quality of the generated frames and the pose generalization performance of our method, testing is specifically performed on novel poses that are not included in the training dataset. It is important to note that due to the stochasticity introduced by factors such as clothing wrinkles, for the unseen poses, ground-truth images represent only \emph{one possible scenario}. Therefore, metrics emphasizing pixel-level correspondence, such as PSNR, may not comprehensively evaluate the fidelity of the generated images. In our evaluation, we primarily employ LPIPS and FID, metrics describing perceptual similarity, while we also provide experimental results with PSNR (in light font).

\textbf{Quantitative Results.} When performing novel pose synthesis on different IDs, we search for the best $t_{split}, k$ pair for each ID regarding their various clothes wrinkles and action dynamics. Tab. \ref{tab:single_frame} presents a quantitative comparison among ARAH, PoseVocab and our method across four sequences of the ZJU-MoCap dataset, we report the average values of these metrics across all test frames. It can be illustrated that our method achieves the best or second-best results across the four sequences. While ARAH achieves the highest PSNR, our approach offers a balanced performance, excelling in LPIPS and FID, which indicates that our method has good generalizability and generative capability given novel poses.

\textbf{Qualitative Results.} The results of the qualitative experiments are depicted in Fig. \ref{fig:single_frame}. ARAH tends to produce relatively blurry images in these scenarios. In contrast, images generated by PoseVocab exhibit relatively clear texture details, albeit with some issues. For instance, the stripes on the T-shirt appear somewhat blurry, and there is color ``bleeding'' from the green clothing onto the arms. Contrarily, the images produced by our method show clearer finer details and a higher sense of realism in clothing wrinkle details.

\subsection{Continuous Video Synthesis}

Fig. \ref{fig:abl_sampling_strategy} illustrates the experimental results of generating a sequence of consecutive frames using our method, which is also performed on novel poses. When the Gaussian consistency sampling strategy is not utilized, and instead, sampling begins with pure Gaussian noise for each frame, the resulting frames experience pronounced inter-frame inconsistency. On the contrary, deriving the 3D Gaussians for the current frame with the Gaussian local coordinates from the previous frame first and then proceeding with subsequent 2D denoising processes significantly enhances the continuity between the output frames.

\begin{table}[t]
    \centering
    \begin{tabular}{@{}c@{\hspace{2pt}}c@{\hspace{3pt}}c@{\hspace{3pt}}c@{\hspace{3pt}}c@{\hspace{3pt}}c@{\hspace{3pt}}c@{}}
    \toprule
    
    \multirow{2}{*}{Method} & \multicolumn{3}{c}{313}  & \multicolumn{3}{c}{315}
    
    \\ \cmidrule(l){2-7}
    & {\textcolor{mygray}{\fontsize{9pt}{9pt}\selectfont PSNR$\uparrow$}} & {\fontsize{9pt}{9pt}\selectfont LPIPS$\downarrow$} & {\fontsize{9pt}{9pt}\selectfont FID$\downarrow$} & {\textcolor{mygray}{\fontsize{9pt}{9pt}\selectfont PSNR$\uparrow$}} & {\fontsize{9pt}{9pt}\selectfont LPIPS$\downarrow$} & {\fontsize{9pt}{9pt}\selectfont FID$\downarrow$}
    
    \\ \midrule
    
    Ours-2D & \textcolor{mygray}{\fontsize{9pt}{9pt}\selectfont 22.7} & \underline{\fontsize{9pt}{9pt}\selectfont 0.086} & \textbf{\fontsize{9pt}{9pt}\selectfont 16.8} & \textcolor{mygray}{\fontsize{9pt}{9pt}\selectfont 19.7} & {\underline{\fontsize{9pt}{9pt}\selectfont 0.098}} & \textbf{\fontsize{9pt}{9pt}\selectfont 19.8} \\
    
    Ours-3D & \textcolor{mygray}{\textbf{\fontsize{9pt}{9pt}\selectfont 24.2}} & \fontsize{9pt}{9pt}\selectfont 0.148 & \fontsize{9pt}{9pt}\selectfont 115.8& \textcolor{mygray}{\textbf{\fontsize{9pt}{9pt}\selectfont 21.6}} & \fontsize{9pt}{9pt}\selectfont 0.150 & \fontsize{9pt}{9pt}\selectfont 79.7\\
    
    Ours-overall & \underline{\textcolor{mygray}{\fontsize{9pt}{9pt}\selectfont 23.5}} & {\textbf{\fontsize{9pt}{9pt}\selectfont 0.080}} & {\underline{\fontsize{9pt}{9pt}\selectfont 19.5}} & \underline{\textcolor{mygray}{\fontsize{9pt}{9pt}\selectfont 20.7}} & \textbf{\fontsize{9pt}{9pt}\selectfont 0.090} & {\underline{\fontsize{9pt}{9pt}\selectfont 20.2}}
    \\ \midrule
    \end{tabular}
    \caption{Quantitative ablation study on our 3D-aware denoising process. Ours-2D uses only 2D denoisers, while Ours-3D retains the initial 3D rectifier but omits later 2D or 3D steps.}
    \label{tab:abl_2d_3d_overall}
\end{table}

\begin{figure}[t]
  \centering
  \includegraphics[width=1.0\columnwidth]{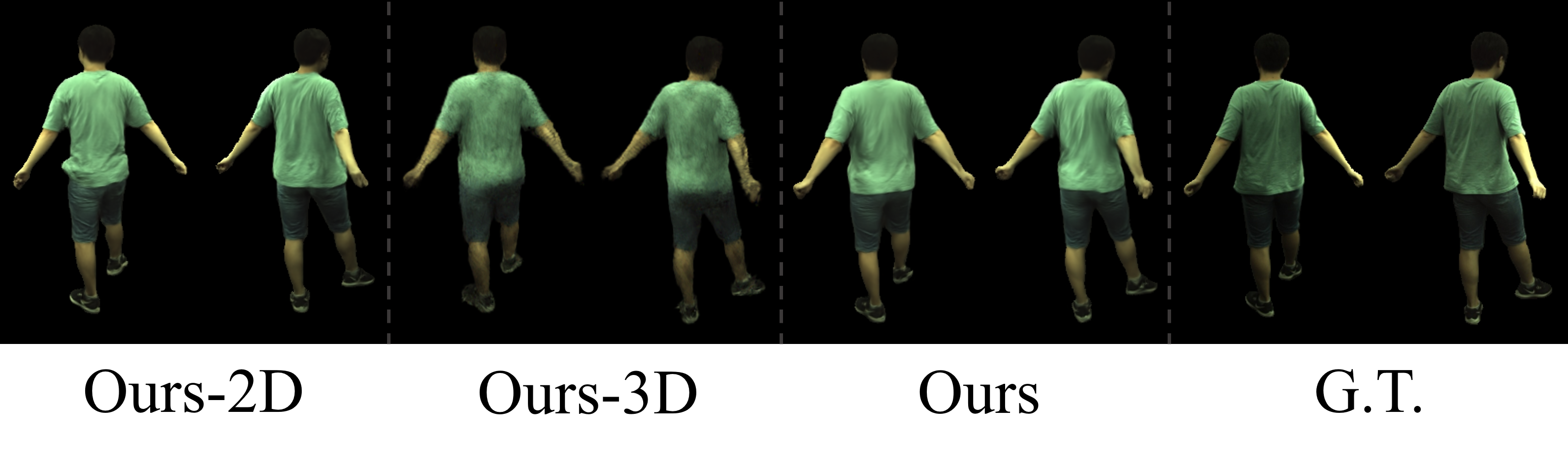}
  \caption{Novel pose synthesis results with different designs of our 3D-aware denoiser.}
  \label{fig:abl_2d_3d_overall}
\end{figure}

\subsection{Ablation Studies}

\textbf{3D-aware Denoising Process.} The complete 3D-aware denoising process consists of two submodules: 3D rectifier and 2D denoiser. LPIPS and FID metrics in Tab. \ref{tab:abl_2d_3d_overall} indicates that conducting only the 2D process yields relatively better performance, while images obtained solely through the 3D counterpart are the least satisfactory. The results in Fig. \ref{fig:abl_2d_3d_overall} reveal that images generated solely through 2D denoising display richer local details but have limitations in overall modeling. In contrast, the 3D process excels in producing reasonable global attributes but fails to model wrinkles and occlusions. Consecutively performing these two processes allows for a synergistic combination of their strengths, as depicted in ``ours''. We further analyze the impact of varying split point $t_{split}$ and the insertion counts of 3D rectifier $k$ on the result. As presented in Tab. \ref{tab:abl_t_split_insert_count}, as $t_{split}$ increases, the 2D denoiser applies stronger corrections, resulting in lower LPIPS and FID values, indicating improved image realism. Moreover, the number of 3D rectifying steps has little impact on image quality when $t_{split}$ is fixed.

\begin{table}[t]
    \centering
    \begin{tabular}{@{}c@{\hspace{1.5pt}}c@{\hspace{1.5pt}}c@{\hspace{0pt}}c@{\hspace{0pt}}c@{\hspace{0pt}}c@{\hspace{0pt}}c@{\hspace{0pt}}c@{\hspace{0pt}}c@{\hspace{0.5pt}}c@{}}
    \toprule
    \multirow{2}{*}{$t_{split}$} & \multirow{2}{*}{$k$} & \multicolumn{2}{c}{313}  & \multicolumn{2}{c}{315}  & \multicolumn{2}{c}{377} & \multicolumn{2}{c}{386}
    \\ \cmidrule(l){3-10}
    & & \fontsize{9pt}{9pt}\selectfont LPIPS$\downarrow$ & \fontsize{9pt}{9pt}\selectfont FID$\downarrow$ & \fontsize{9pt}{9pt}\selectfont LPIPS$\downarrow$ & \fontsize{9pt}{9pt}\selectfont FID$\downarrow$ & \fontsize{9pt}{9pt}\selectfont LPIPS$\downarrow$ & \fontsize{9pt}{9pt}\selectfont FID$\downarrow$ & \fontsize{9pt}{9pt}\selectfont LPIPS$\downarrow$ & \fontsize{9pt}{9pt}\selectfont FID$\downarrow$
    \\ \midrule
    \phantom{200} & 2 & \fontsize{9pt}{9pt}\selectfont 0.082 & \fontsize{9pt}{9pt}\selectfont 21.5 & \fontsize{9pt}{9pt}\selectfont 0.093 & \fontsize{9pt}{9pt}\selectfont 21.9 & \fontsize{9pt}{9pt}\selectfont 0.094 & \fontsize{9pt}{9pt}\selectfont 29.4 & \fontsize{9pt}{9pt}\selectfont 0.127 & \fontsize{9pt}{9pt}\selectfont 31.8 \\
    200 & 3 & \fontsize{9pt}{9pt}\selectfont 0.082 & \fontsize{9pt}{9pt}\selectfont 23.1 & \fontsize{9pt}{9pt}\selectfont 0.093 & \fontsize{9pt}{9pt}\selectfont 21.7 & \fontsize{9pt}{9pt}\selectfont 0.093 & \fontsize{9pt}{9pt}\selectfont 28.6 & \fontsize{9pt}{9pt}\selectfont 0.127 & \fontsize{9pt}{9pt}\selectfont 31.9 \\
    \phantom{200} & 4 & \fontsize{9pt}{9pt}\selectfont 0.082 & \fontsize{9pt}{9pt}\selectfont 21.7 & \fontsize{9pt}{9pt}\selectfont 0.093 & \fontsize{9pt}{9pt}\selectfont 21.4 & \fontsize{9pt}{9pt}\selectfont 0.094 & \fontsize{9pt}{9pt}\selectfont 28.5 & \fontsize{9pt}{9pt}\selectfont 0.127 & \fontsize{9pt}{9pt}\selectfont 32.3 \\ \midrule
    
    \phantom{300} & 2 & \textbf{\fontsize{9pt}{9pt}\selectfont 0.080} & \textbf{\fontsize{9pt}{9pt}\selectfont 19.5} & \fontsize{9pt}{9pt}\selectfont 0.092 & \fontsize{9pt}{9pt}\selectfont 21.2 & \underline{\fontsize{9pt}{9pt}\selectfont 0.091} & \underline{\fontsize{9pt}{9pt}\selectfont 27.2} & \textbf{\fontsize{9pt}{9pt}\selectfont 0.123} & \textbf{\fontsize{9pt}{9pt}\selectfont 28.6} \\
    300 & 3 & \underline{\fontsize{9pt}{9pt}\selectfont 0.081} & \fontsize{9pt}{9pt}\selectfont 20.0 & \underline{\fontsize{9pt}{9pt}\selectfont 0.091} & \underline{\fontsize{9pt}{9pt}\selectfont 20.5} & \textbf{\fontsize{9pt}{9pt}\selectfont 0.090} & \textbf{\fontsize{9pt}{9pt}\selectfont 26.4} & \underline{\fontsize{9pt}{9pt}\selectfont 0.124} & \underline{\fontsize{9pt}{9pt}\selectfont 29.2} \\
    \phantom{300} & 4 & \underline{\fontsize{9pt}{9pt}\selectfont 0.081} & \underline{\fontsize{9pt}{9pt}\selectfont 19.7} & \textbf{\fontsize{9pt}{9pt}\selectfont 0.090} & \textbf{\fontsize{9pt}{9pt}\selectfont 20.2} & \underline{\fontsize{9pt}{9pt}\selectfont 0.091} & \fontsize{9pt}{9pt}\selectfont 27.6 & \textbf{\fontsize{9pt}{9pt}\selectfont 0.123} & \fontsize{9pt}{9pt}\selectfont 29.5 \\ \midrule
    \end{tabular}
    \caption{Comparison of image generation quality under varying $t_{split}$ and $k$.}
    \label{tab:abl_t_split_insert_count}
\end{table}

\begin{figure}[t]
  \centering
  \includegraphics[width=1.0\columnwidth]{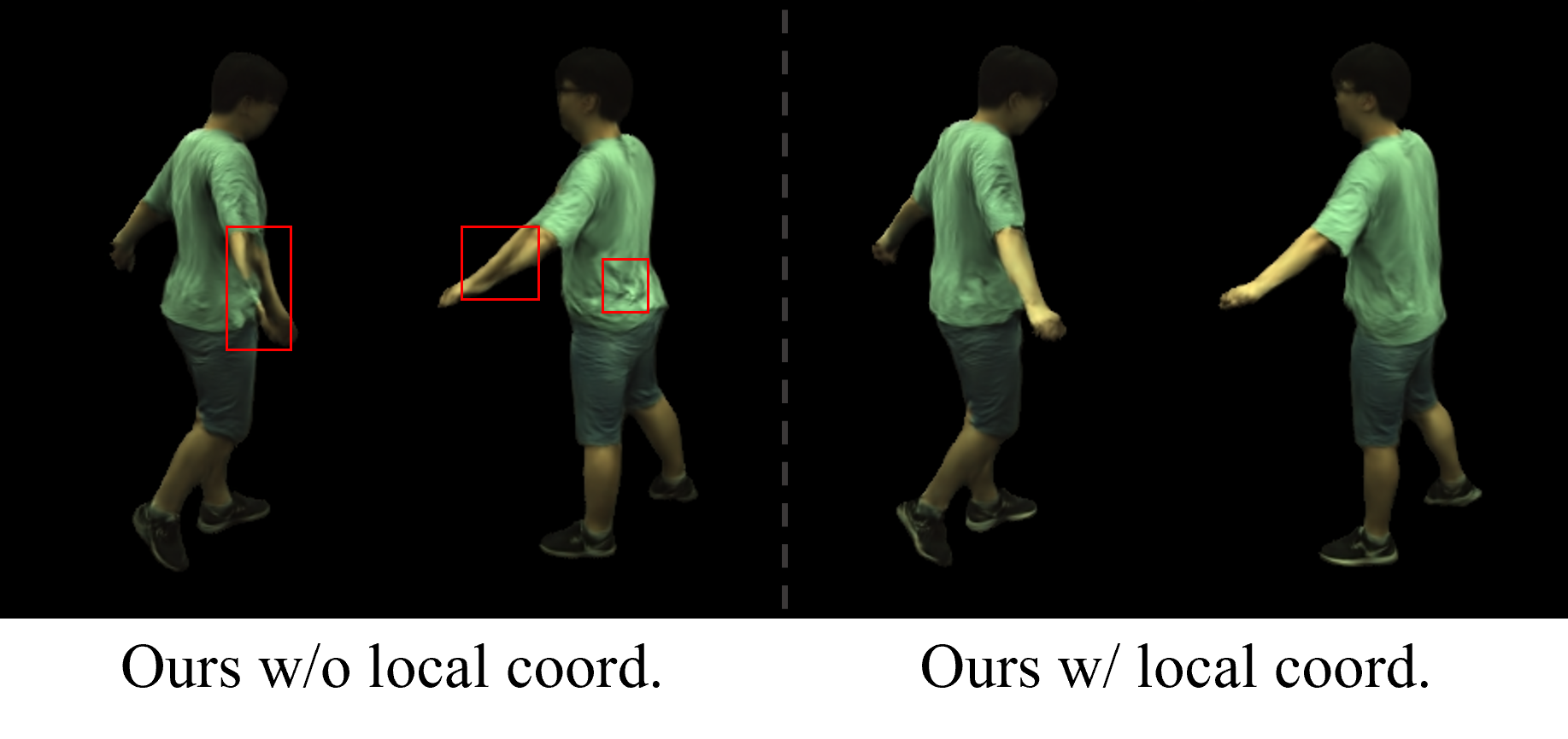}
  \caption{Novel pose synthesis results without (w/o) and with (w/) our mesh-based local coordinate representation.}
  \label{fig:abl_local_coord}
\end{figure}

\textbf{Gaussian Local Coordinate Representation.} We conduct an ablation study comparing the image generation quality of our framework with and without our mesh-based Gaussian local coordinate representation. From Fig. \ref{fig:abl_local_coord}, we can see that this representation allows 3D Gaussians to move flexibly within a reasonable range, resulting in more realistic modeling of body and clothes details, such as wrinkles and occlusions. In contrast, a 3D-aware denoiser without this representation lacks the ability to express details reasonably. As a consequence, the generated images tend to exhibit unnatural clothing wrinkles and fail to accurately model limbs, leading to a significant decline in visual quality.

\section{Conclusion}

We present 3D$^2$-Actor, an innovative pose-conditioned 3D-aware denoiser designed for the high-fidelity reconstruction and animation of 3D human avatars. Our approach employs a 2D denoiser to refine the intricate details of noised images from the previous step, generating high-quality clean images that facilitate the 3D reconstruction process in the subsequent 3D rectifier. Complementing this, our 3D rectifier employs a two-stage projection strategy with a novel local coordinate representation to render multi-view images with enhanced 3D consistency by incorporating 3DGS-based techniques. Additionally, we introduce a Gaussian consistency sampling strategy that improves inter-frame continuity in video synthesis without additional training overhead. Our method achieves realistic human animation and high-quality dynamic video generation with novel poses.

\section{Acknowledgments}
This work is partly supported by the National Key R\&D Program of China (2022ZD0161902), Beijing Municipal Natural Science Foundation (No. 4222049), the National Natural Science Foundation of China (No. 62202031), and the Fundamental Research Funds for the Central Universities.

\bibliography{aaai25}



\section*{Supplementary material}

# Supplementary Material for 3D$^2$-Actor: Learning Pose-Conditioned 3D-Aware Denoiser for Realistic Gaussian Avatar Modeling

## Network Architectures

### Pixel-Aligned Feature Extraction

The pixel-aligned feature extraction network in our 3D rectifier can be viewed as a U-Net (Ronneberger, Fischer, and Brox 2015) architecture, as depicted in Fig. 1. The network takes clean images and body segmentation maps as inputs, finally producing pixel-aligned feature maps. Initially, a 2D convolution is applied to the input image to expand its channels to 16. Following that, a ResNet (He et al. 2016) block is employed to produce intermediate feature maps. Next, the feature maps are modulated by an SFT (Wang et al. 2018) layer based on the segmentation map. Following the last ResNet block in the current U-Net layer, the output is downsampled and forwarded to the subsequent layer. The network comprises a total of three layers, with intermediate feature dimensions of 32, 64, and 128, respectively. Additionally, an extra self-attention (Vaswani 2017) module is introduced in the middle of the U-Net.

### 2D Denoiser

Our 2D denoiser shares a similar structure to the pixel-aligned feature extraction network, with the distinction of extending it to four layers. It takes noised images, body segmentation maps and the denoising timestep as input, finally predicting clean images. The denoising timestep is processed by a SiLU (Hendrycks and Gimpel 2016) activation function and a linear block, then injected into the Scale & Shift layer (refer to Fig. 1). The number of the intermediate feature channels in each layer is 8, 16, 32, and 64, respectively. Finally, a 1×1 convolution is applied to convert the channel number to 3, corresponding to the RGB channels.

## Additional Experiments

To ensure completeness of our experiments, we conducted several additional experiments on the ZJU-MoCap (Peng et al. 2021) dataset, which were not included in the main paper. All the reported metrics are consistent with those used in the main text, including Learned Perceptual Image Patch Similarity (LPIPS) (Zhang et al. 2018), Frechet Inception Distance (FID) (Heusel et al. 2017), and Peak Signal-to-Noise Ratio (PSNR, indicated in light font).

### Design of Our 3D-aware Denoising Process

We present supplementary ablation results on the design of our 3D-aware denoiser, with experiments conducted on sequences "377" and "386". The results are depicted in Tab. 1 and Fig. 2. Without employing our 3D rectifier, the images produced by a 3D-aware denoiser filled with 2D denoising steps exhibit better image quality but lack multi-view consistency (as indicated by red bounding boxes). If we merely employ one 3D rectifying step and cancel the later 2D or 3D steps, the generated images will be good in overall 3D structure but may demonstrate some blurriness and artifacts. A 3D-aware denoiser that effectively integrates 3D and 2D steps can produce images with both high 3D consistency and visual quality.

| Method | 377 | | | 386 | | |
|---|---|---|---|---|---|---|
| | PSNR↑ | LPIPS↓ | FID↓ | PSNR↑ | LPIPS↓ | FID↓ |
| Ours-2D | 23.5 | 0.098 | **19.8** | 26.2 | **0.121** | **25.4** |
| Ours-3D | 24.0 | 0.180 | 158.4 | 26.6 | 0.250 | 138.8 |
| Ours-overall | **24.4** | **0.090** | 26.4 | **26.8** | 0.123 | 28.6 |

Table 1: Quantitative ablation study on our 3D-aware denoising process. Ours-2D uses only 2D denoisers, while Ours-3D retains the initial 3D rectifier but omits later 2D or 3D steps.

We also provide additional results on the choice of different split times $t_{split}$ and insertion counts $k$ of the 3D rectifier in our 3D-aware denoising process. The results are shown in Tab. 2. When $t_{split} = 300$, the result images achieve relatively better visual quality (lower FID and LPIPS scores). In addition, since the motion dynamics and the complexity of clothes textures varies a lot across the four sequences of human actors, the optimal value of $k$ is slightly different for various sequences. Note that results are not presented here for $t_{split} > 300$, as the PSNR value is not ideal enough under such circumstance. The rationale is that an increase in $t_{split}$ leads to the addition of more 2D denoising steps at the late stage of the 3D-aware denoising process, which inevitably compromises the 3D structure of the human avatar.

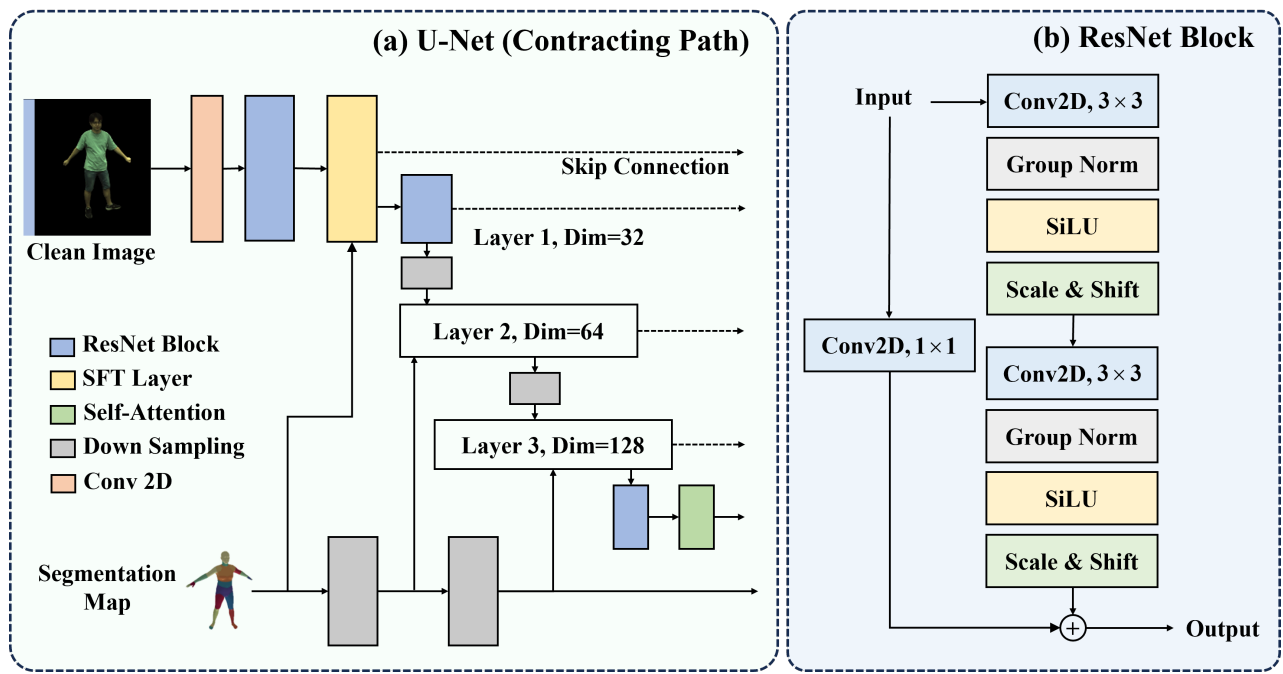

Figure 1: Illustration of our pixel-aligned feature extraction network. (a) Only the contracting path of the U-Net is displayed. The expansive path follows a similar architecture but replaces downsampling with the upsampling process. (b) A detailed view of our ResNet Block.

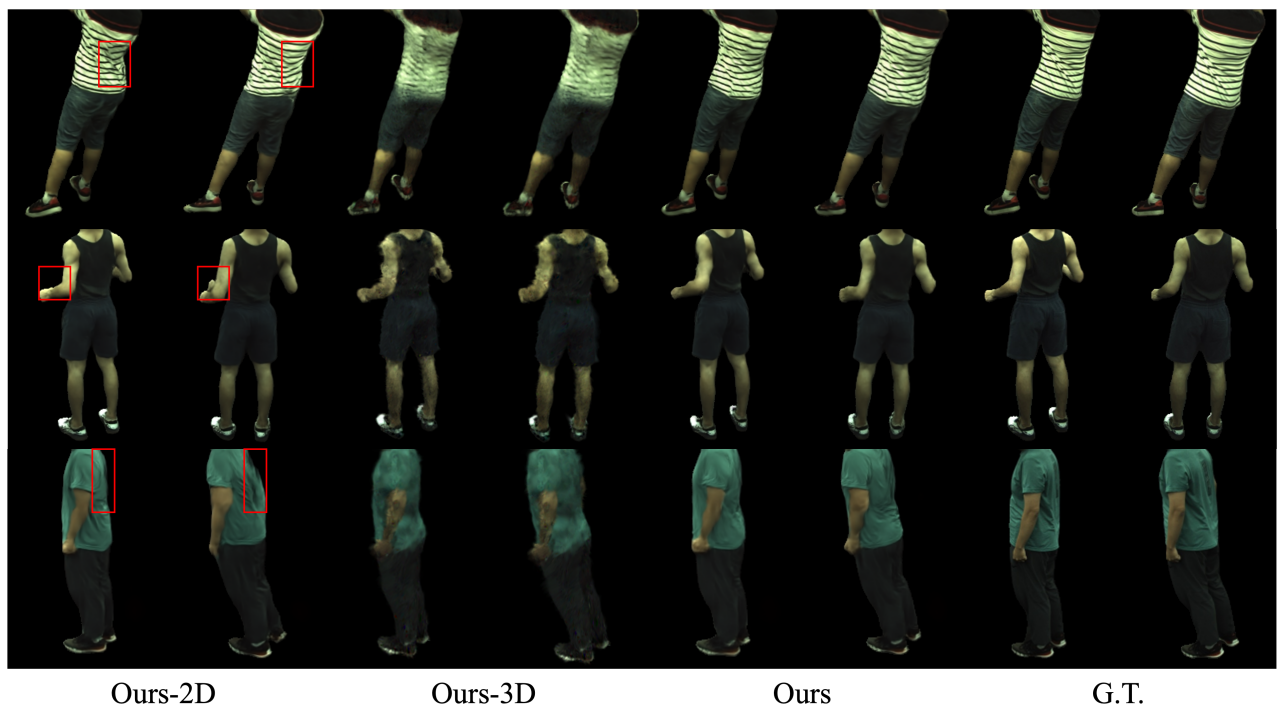

Figure 2: Additional novel view synthesis results generated with different design of our 3D-aware denoising process. "Ours-2D" utilizes solely 2D denoising steps, while "Ours-3D" retains the initial 3D rectifying step but excludes subsequent 2D or 3D steps. "Ours" denotes our final design. Row 1 to 3 present the results from sequence "315", "377" and "386", respectively. For each sequence, we display the generated images from two different camera views.

| $t_{split}$ | $k$ | 313 | | 315 | | 377 | | 386 | |
|---|---|---|---|---|---|---|---|---|---|
| | | LPIPS↓ | FID↓ | LPIPS↓ | FID↓ | LPIPS↓ | FID↓ | LPIPS↓ | FID↓ |
| 100 | 2 | 0.086 | 25.6 | 0.096 | 24.1 | 0.100 | 34.5 | 0.134 | 37.4 |
| 100 | 3 | 0.086 | 24.6 | 0.096 | 23.1 | 0.099 | 32.9 | 0.134 | 38.5 |
| 100 | 4 | 0.086 | 24.8 | 0.096 | 23.3 | 0.099 | 32.8 | 0.134 | 37.4 |
| 100 | 5 | 0.085 | 25.3 | 0.096 | 23.3 | 0.099 | 33.0 | 0.135 | 38.4 |
| 200 | 2 | 0.082 | 21.5 | 0.093 | 21.9 | 0.094 | 29.4 | 0.127 | 31.8 |
| 200 | 3 | 0.082 | 23.1 | 0.093 | 21.7 | 0.093 | 28.6 | 0.127 | 31.9 |
| 200 | 4 | 0.082 | 21.7 | 0.093 | 21.4 | 0.094 | 28.5 | 0.127 | 32.3 |
| 200 | 5 | 0.081 | 21.5 | 0.092 | 20.9 | 0.093 | 28.7 | 0.127 | 32.3 |
| 300 | 2 | **0.080** | **19.5** | 0.092 | 21.2 | <u>0.091</u> | <u>27.2</u> | **0.123** | **28.6** |
| 300 | 3 | <u>0.081</u> | 20.0 | <u>0.091</u> | <u>20.5</u> | **0.090** | **26.4** | <u>0.124</u> | 29.2 |
| 300 | 4 | <u>0.081</u> | <u>19.7</u> | **0.090** | **20.2** | <u>0.091</u> | 27.6 | **0.123** | 29.5 |
| 300 | 5 | <u>0.081</u> | <u>19.7</u> | 0.092 | 21.4 | <u>0.091</u> | 27.7 | **0.123** | 29.5 |

Table 2: Comparison of image generation quality under varying $t_{split}$ and $k$.

### Inter-frame Gaussian Consistency Sampling

We provide animated avatar videos under novel poses in our supplementary materials, generated with and without our proposed Inter-frame Gaussian Consistency (IGC) Sampling strategy. These videos demonstrate that our IGC sampling strategy significantly enhances inter-frame continuity and reduces flickering in the generated animations. For a quantitative ablation study, we generate two 100-frame videos with (w/) and without (w/o) the IGC sampling for each of the four IDs and calculated the average LPIPS and optical flow warp error between adjacent frames. A lower value indicates better inter-frame continuity in the generated video. As shown in Tab. 3 and Tab. 4, the metric values are considerably lower when employing the IGC sampling, thereby validating its effectiveness.

| **Strategy** | 313 | 315 | 377 | 386 |
|---|---|---|---|---|
| w/o IGC sampling | 0.057 | 0.077 | 0.078 | 0.091 |
| w/ IGC sampling | **0.045** | **0.048** | **0.071** | **0.085** |

Table 3: Average LPIPS values between adjacent frames with and without the IGC sampling strategy.

| **Strategy** | 313 | 315 | 377 | 386 |
|---|---|---|---|---|
| w/o IGC sampling | 1.84 | 2.00 | 1.05 | 0.67 |
| w/ IGC sampling | **1.41** | **1.06** | **0.95** | **0.58** |

Table 4: Average optical flow warp error between adjacent frames with and without the IGC sampling strategy.

### Mesh-based Local Coordinate Representation

Additional qualitative comparisons of the generated images with and without our mesh-based Gaussian local coordinate representation are presented in Fig. 3. The experiments were conducted on sequences "315", "377" and "386". It is evident that a 3D-aware denoising process without the local coordinate representation falls short in producing images that exhibit natural-looking clothes wrinkles and body appearances, leading to a marked degradation in overall visual quality. Conversely, with the assistance of this representation, the generated images boast superior visual fidelity.

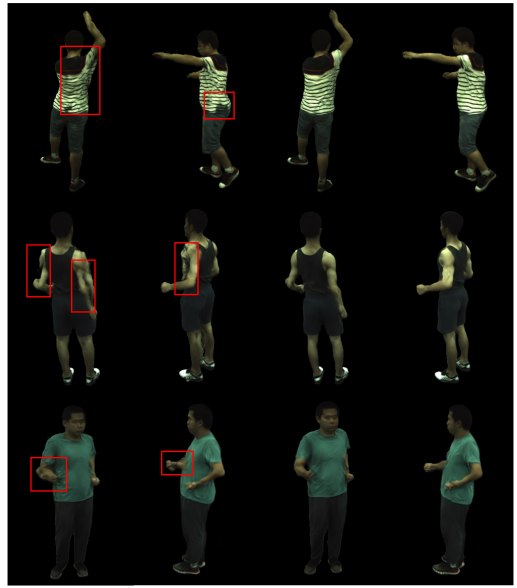

Ours w/o local coord.    Ours w/ local coord.

Figure 3: Additional novel view synthesis results generated without (w/o) and with (w/) our mesh-based local coordinate representation. Row 1 to row 3 present the results from sequence "315", "377" and "386", respectively.

### 3D Multi-view Consistency

To demonstrate 3D multi-view consistency in novel views, we also include a video rendered with smooth and wide-ranging camera movement in the supplementary materials.



\end{document}